\documentclass[letterpaper]{article} % DO NOT CHANGE THIS
\usepackage{aaai2026}  % DO NOT CHANGE THIS
\usepackage{times}  % DO NOT CHANGE THIS
\usepackage{helvet}  % DO NOT CHANGE THIS
\usepackage{courier}  % DO NOT CHANGE THIS
\usepackage[hyphens]{url}  % DO NOT CHANGE THIS
\usepackage{graphicx} % DO NOT CHANGE THIS
\usepackage{natbib}  % DO NOT CHANGE THIS AND DO NOT ADD ANY OPTIONS TO IT
\usepackage{caption} % DO NOT CHANGE THIS AND DO NOT ADD ANY OPTIONS TO IT
\usepackage{amsfonts} % For \mathbb
\usepackage{amsmath}
\usepackage{algorithm}
\usepackage{algorithmic}

\usepackage{amsthm} % 定理环境支持
\newtheorem{theorem}{Theorem} % 创建定理环境

\usepackage{multirow} 
\usepackage{booktabs}

\usepackage{newfloat}
\usepackage{listings}
\DeclareCaptionStyle{ruled}{labelfont=normalfont,labelsep=colon,strut=off} % DO NOT CHANGE THIS
\floatstyle{ruled}
\newfloat{listing}{tb}{lst}{}
\floatname{listing}{Listing}
\title{Perturbation-mitigated USV Navigation with Distributionally Robust Reinforcement Learning}
\author{
    Zhaofan Zhang\textsuperscript{\rm 1},
    Minghao Yang\textsuperscript{\rm 1},
    Sihong Xie\textsuperscript{\rm 1}\thanks{Corresponding authors.},
    Hui Xiong\textsuperscript{\rm 1,2}\footnotemark[1]
}
\affiliations{
    \textsuperscript{\rm 1}The Hong Kong University of Science and Technology (Guangzhou)\\
    \textsuperscript{\rm 2}The Hong Kong University of Science and Technology\\
}

\usepackage{bibentry}
\begin{document}

\maketitle

\begin{abstract}
The robustness of Unmanned Surface Vehicles (USV) is crucial when facing unknown and complex marine environments, especially when heteroscedastic observational noise poses significant challenges to sensor-based navigation tasks. Recently, Distributional Reinforcement Learning (DistRL) has shown promising results in some challenging autonomous navigation tasks without prior environmental information. However, these methods overlook situations where noise patterns vary across different environmental conditions, hindering safe navigation and disrupting the learning of value functions. 
% To address the problem, we propose DRIQN, a novel distributionally robust optimization (DRO) enhanced DistRL framework with implicit quantile networks, which prioritizes worst-case performance of natural conditions while incorporating heterogeneous noise types through explicit subgroup modeling in the replay buffer.
To address the problem, we propose DRIQN to integrate Distributionally Robust Optimization (DRO) with implicit quantile networks to optimize worst-case performance under natural environmental conditions. Leveraging explicit subgroup modeling in the replay buffer, DRIQN incorporates heterogeneous noise sources and target robustness-critical scenarios.
% This framework enhances the robustness of the planner by focusing on the worst-performing noise subgroups, particularly in environments with complex noise patterns. 
Experimental results based on the risk-sensitive environment demonstrate that DRIQN significantly outperforms
state-of-the-art methods, achieving +13.51\% success rate, -12.28\% collision rate and +35.46\% for time saving, +27.99\% for energy
saving, compared with the runner-up.

% validate the superiority of DRIQN over state-of-the-art methods in achieving USV navigation under observational perturbation across various performance metrics.

% demonstrate that compared to other state-of-the-art methods, the proposed DRIQN achieves superior success rates, improved cumulative rewards, and reduced collision frequencies across diverse perceptual noise patterns.
\end{abstract}


\section{Introduction}
% figure 1
\begin{figure}[t]
\centering
% \vspace{0.21cm}
\includegraphics[width=0.98\linewidth]{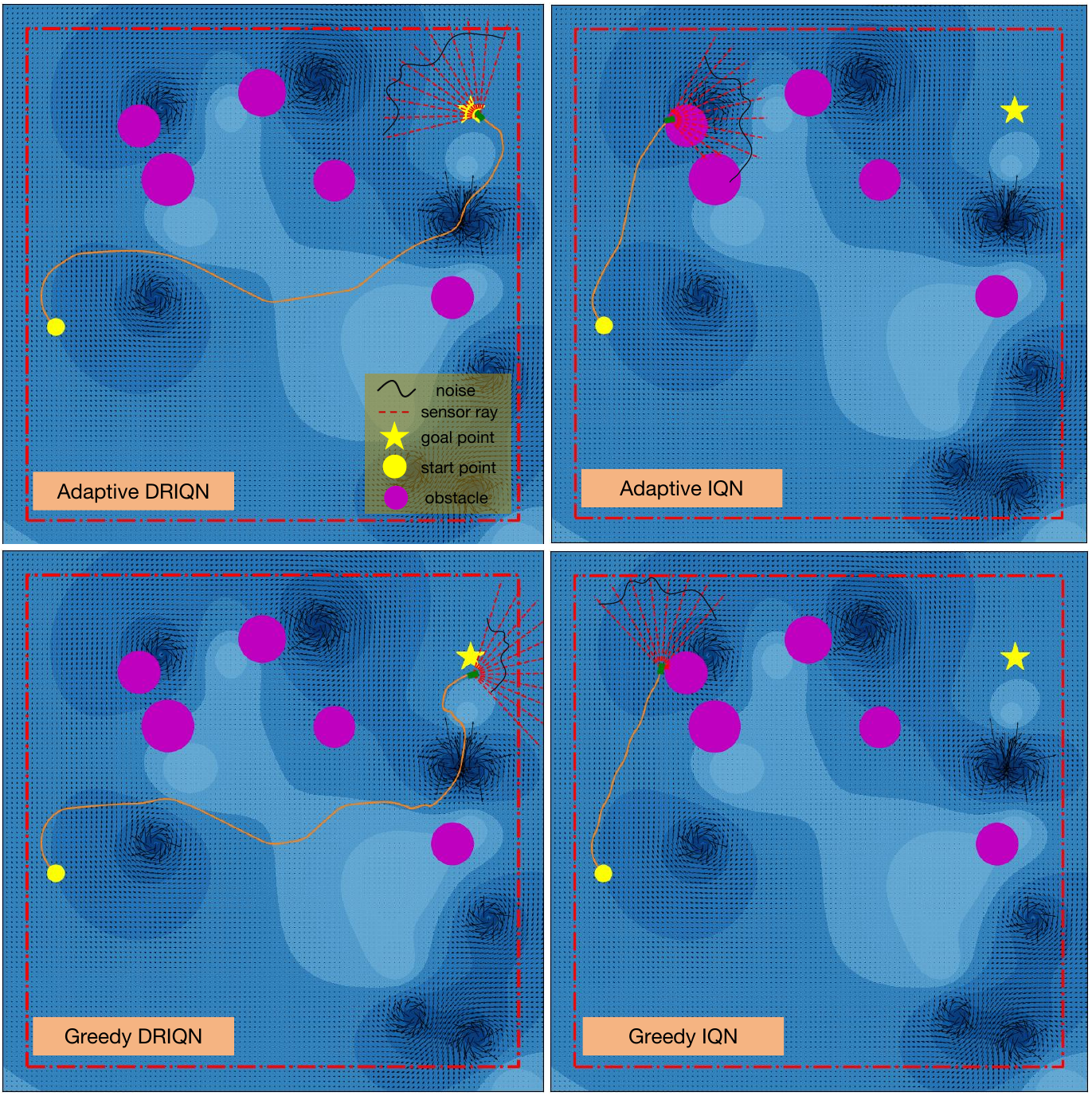}
\caption{\textbf{Proposed DRIQN \textsc{vs.} IQN: }Policy trajectories visualization with observational noise from sensor. For IQN-based methods, we investigate adaptive strategy and greedy strategy with details in Section \emph{Metrics and Strategy}.}
\label{fig1}
\end{figure}

% perception error due to noise 
%%%%%%%%%%%%%%%

%%%%%%%%%%%%%%%
Autonomous navigation is the keystone capability that transforms an USV from a remotely operated craft into a truly independent maritime agent. Reliable navigation empowers USVs to undertake long-duration missions ranging from port oil spill cleanup \cite{elmakis2023usv} and environmental monitoring \cite{tran2024unmanned} to border security \cite{fernandes2025smart} and search-and-rescue \cite{wang2023cooperative} while simultaneously reducing operational risk, human exposure, and overall cost. In these maritime navigation tasks, the environment is characterized by complexity and partial observability. The intricate variations in water currents are difficult to monitor in real-time, and obstacles such as reefs may also be encountered \cite{lolla2015path}. Moreover, the perception devices used by the robot may face various unobservable interferences within the environment, resulting in noisy observations \cite{vieira2020underwater}.
Specifically, heterogeneous extreme natural conditions (e.g., weather, sea state, geographic locality) induce trajectory-specific, non-stationary observational noise. For instance, as illustrated in Figure~\ref{fig1}, the observational perturbation exacerbates the inherent uncertainty of complex marine environments and impedes accurate risk-sensitive policy learning based on corrupted observation, creating substantial navigation hazards for USVs.
Prior works on USV navigation can be mainly grouped into: (i) \emph{model-based} methods \cite{wang2023efficient, menges2024nonlinear, asfihani2025fault}; (ii) models with \emph{prior-dependent} knowledge \cite{han2016gps, hong2024model}; (iii) \emph{direct observation-driven decision-making} approaches (e.g., APF \cite{Sun2019SmartOA}, BA \cite{McGuire2018ACS} and reinforcement learning \cite{wang2022sim, lin2023robust}). Direct observation-driven approaches make decisions based on sensor observation without prior environmental knowledge and extra modeling design. 
Despite promising results, model-based approaches (e.g., nonlinear MPC) incorporate analytical vehicle dynamics into optimal control, requiring high-fidelity models and suffering degradation under uncertainties. Prior-dependent methods further face limitations in prior availability and freshness, incurring substantial precomputation overhead. In contrast, observation-driven decision-making methods offer a generalizable, prior-independent solution for navigation in unknown environments, albeit with heightened sensitivity to observational noise.
Recently, as one kind of prominent observation-driven decision-making method, Deep Reinforcement Learning (DRL) has emerged for robotic navigation in complex environments \cite{lee2023adaptive,hossain2024quasinav}.
% demonstrating performance advantages over model-based or prior-dependent approaches. 
% DRL learns adaptive navigation policies by maximizing the state-action value function through environmental interaction, replacing the need for difficult-to-obtain prior knowledge with sensor observations \cite{wang2024research}. DistRL extends standard RL by modeling the distribution of the value function rather than merely its expectation via quantiles \cite{liang2024bridging}. 
Building on this advantage, some DistRL methods~\cite{lin2023robust,liu2023adaptive} enable risk-sensitive navigation through quantile-based return distribution modeling. By employing task-specific reward designs and distortion functions with Conditional Value-at-Risk (CVaR), DistRL reshapes return distributions to focus on critical lower-tail outcomes while constraining distributional support, thereby enhancing robust policy learning.
% Building on pioneering works C51 \cite{bellemare2017distributional} and QRDQN \cite{dabney2018distributional}, the Implicit Quantile Network (IQN) \cite{dabney2018implicit} enables more flexible distribution modeling through quantile sampling and. IQN-based methods [9,10] leverage distributional properties (e.g., lower tail variance) to quantify risk and adjust distortion functions, constraining the support of the return distributions.
% However, regional transitions and diverse extreme natural conditions induce trajectory‑dependent, heterogeneous observation noise, whose nonstationary patterns impede accurate risk-sensitive policy learning based on the state–action value distribution and severely degrade both safety and task success rates. This work focuses on investigating the performance of traditional methods and RL-based works in these grueling conditions and how to improve the robustness of DistRL for safe navigation in environments with region-dependent noise patterns.
% Nevertheless, these methods fail to account for the safety with corrupted observations caused by heterogeneous noisy conditions.
Nevertheless, without environmental priors or modeling knowledge for constraint factors, these approaches cannot be readily optimized via constructing robust mechanisms, thereby lacking explicit safety guarantees under heterogeneous noise-corrupted observations.
% However, these methods overlook the heterogeneous extreme natural conditions (e.g., weather, sea state, geographic locality) inducing trajectory-specific, non-stationary observational noise that corrupt the sensors. 
% For instance, as illustrated in Figure~\ref{fig1}, the observational noise impedes accurate risk-sensitive policy learning based on state-action value distributions, severely degrading safety and success. 
This presents a pressing research challenge: how can we ensure the safety of unmanned surface navigation in multi-noise-corrupted marine environment just using a data-driven optimization approach?

To address the challenge, we propose Distributionally Robust Implicit Quantile Networks (DRIQN), a novel framework ensuring risk-sensitive policies via quantile regression under observational noise. Leveraging distributionally robust optimization, DRIQN reformulates the DRO problem as a semi-definite program (Eq.~\ref{eq:quadratic program}) to holistically address diverse noise influences.
Crucially, DRIQN implements ``gradient substitution'' replacing the network's output-layer gradients with DRO-based gradients to enhance robustness without auxiliary modules.
Considering the scalability challenge of DRO caused by the large-scale datapoint-level optimization process, we transform standard datapoint-based DRO by leveraging transition data's inherent structure: under specific environmental conditions, noisy transitions naturally form homogeneous subgroups following corresponding noise patterns. This structural property enables reformulating the original datapoint-level DRO problem as the supremum over subgroup-level loss functions. Each subgroup represents unlabeled transition data clusters with distinct noise characteristics.
% Further more, as one optimization-based method, it's hard to directly apply DRO to big data scenarios and large parameterized models \cite{shen2022towards}.
% For this problem, transition data exhibiting distinct noise patterns within specific environmental conditions form homogeneous subgroups. This character allows reformulating the original datapoint-DRO problem as the supremum over subgroup-level loss functions. Thus, we focus on the uncertainty set $\mathcal{Q}$, which is partitioned into \(J\) subgroups that represent the noise-label agnostic subgroups of transition data with different noise types.
Experimental results demonstrate that DRIQN significantly outperforms baselines: achieving +13.51\% success rate, -12.28\% collision rate and +35.46\% for time saving, +27.99\% for energy saving, compared with the runner-up. 
To our knowledge, DRIQN is the first method tackling multiple observational noise patterns from natural conditions, enabling robust risk-sensitive policies for USV navigation.

The main contributions of this paper are as follows:
\begin{itemize}
    % \item We introduce a novel data-driven optimization method suitable for distributional reinforcement learning that leverages DRO to compute and replace gradients of neural networks, significantly improving robustness while ensuring efficiency and broad applicability.
    % \item We propose a novel and neat distributionally robust optimization framework that computes and replaces the gradients of neural networks to improve the RL agent's performance against perturbed observations.
    \item We propose a novel, streamlined distributionally robust optimization framework that enhances RL agent performance against perturbed observations through substitution of DRO-based gradients.
    \item We present a risk-sensitive and robust navigation approach for developing an online path planner by unifying DistRL and DRO, maintaining robustness and safety in the presence of complex environmental noise patterns.
    \item Simulated experiments show the resulted planner enabled the vehicle to resist the impact of observational perturbation and reach the goal as safely as possible under strong disturbances, unknown flows and obstacles.
\end{itemize}

\section{Related Work}

% \subsection{Robot Safe Navigation Methods}
% Previously, Artificial Potential Field (APF) method [13] navigates robots via goal-attractive and obstacle-repulsive forces, yet suffers from local minima and goal-proximity failures. Subsequent improvements introduced virtual targets [14], annealing techniques [15], dynamic windows [16], and enhanced repulsion [17,18], while other works enhanced dynamic environment adaptability through velocity-aware potential functions [19] or evolutionary algorithms [20,21]. Separately, Bug Algorithms (e.g., Bug1/Bug2 [23]) utilize boundary-following behaviors for reactive planning, later extended to wheeled robots [24], quadrotors [25], and USVs [12]. Enhanced variants (Alg1/Alg2 [26], Rev1/Rev2 [27]) optimize path efficiency by adaptive obstacle-following, and sensor-based methods (VisBug [28,29], TangentBug [30]) leverage range sensing for shorter paths [12].
% Existing works on studying USV navigation fall into three categories: model-based, prior-dependent, and direct observation-driven decision-making approaches.
% Compared with other kinds of methods, direct observation-driven decision-making methods provide a general and low-prior-cost way for path planning.
% Existing research on USV navigation broadly categorizes approaches into three paradigms: model-based, prior-dependent, and direct observation-driven decision-making. 
Among three paradigms (model-based, prior-dependent, and direct observation-driven decision-making) in USV navigation, observation-driven methods offer enhanced generality and reduced prior knowledge requirements for path planning in unknown environments.
As a foundational observation-driven navigation approach, Artificial Potential Field (APF) methods \cite{Khatib1985RealTimeOA, Sun2019SmartOA} navigate robots via goal-attractive and obstacle-repulsive forces, yet suffer from local minima and goal-proximity failures. 
% Subsequent improvements introduced virtual targets [14], annealing techniques [15], dynamic windows [16], and enhanced repulsion [17,18], while other works enhanced dynamic environment adaptability through velocity-aware potential functions [19] or evolutionary algorithms [20,21]. 
Separately, Bug Algorithms \cite{Lumelsky1986DynamicPP} and recent variants Bug1/Bug2 \cite{McGuire2018ACS} utilize boundary-following for reactive planning but generate highly suboptimal paths due to exhaustive obstacle perimeter traversal. 
% Enhanced sensor-based variants \cite{Lumelsky1990IncorporatingRS, Kamon1996ANR} leverage range sensing for shorter paths yet remain sensitive to sensor noise and complex obstacle geometries.
Recently, DRL has been widely applied in navigation tasks.
% enabling agents to learn optimal policies for navigating in complex environments including marine surface \cite{mirowski2016learning,dhiman2018critical,goldsztejn2023ptdrl,lee2023adaptive}. 
% Considering safety in navigation task, safe RL methods \cite{Ltjens2018SafeRL, Dawood2024ADS, tomilin2025hasard} enhance navigation safety by incorporating constraints into optimization objectives. However, their effectiveness is limited by the need for handcrafted constraint design and vulnerability to catastrophic failures induced by transient constraint violations. 
% % Due to the balance of efficiency and safety, DistRL methods are increasingly applied in tasks that require risk-sensitive decision-making in unknown environments affected by terrain conditions, water currents, and airflow, such as quadrupedal locomotion \cite{shi2024robust}, pursuit-evasion \cite{zhang2025terl}, USV \cite{lin2023robust}, and UAV \cite{liu2023adaptive} navigation. 
% DistRL methods' balance of efficiency and safety has led to increasing adoption for risk-sensitive decision-making in complex unknown environments. Representative applications include quadrupedal locomotion \cite{shi2024robust}, pursuit-evasion \cite{zhang2025terl}, USV \cite{lin2023robust}, and UAV \cite{liu2023adaptive} navigation.
% However, their performance deteriorates sharply under severe noise and cannot uphold adequate success rates or safety, which in turn poses a substantial barrier to mission execution in noisy environmental conditions.
Considering safety in navigation task, safe RL methods \cite{Ltjens2018SafeRL, Dawood2024ADS, tomilin2025hasard} enforce navigation safety via constrained optimization, yet require handcrafted constraints and remain vulnerable to catastrophic failures during transient violations.
DistRL's efficiency-safety balance promotes adoption in risk-sensitive applications (e.g., quadruped locomotion \cite{shi2024robust}, pursuit-evasion \cite{zhang2025terl}, USV/UAV navigation \cite{lin2023robust, liu2023adaptive}).
However, severe noise induces sharp performance decay, compromising both success rates and safety in noisy missions.

Considering robustness, distributionally robust optimization is an effective approach for data-driven decision-making in the presence of uncertainty \cite{nietert2023outlier}. The DRO problem is a minimax optimization that seeks a decision minimizing the worst-case expected loss within a distributional uncertainty set. 
By defining the uncertainty set based on the empirical data distribution, DRO method learns a model that remains robust to distributional uncertainties of data \cite{staib2019distributionally}. Whereas expected‑value optimization under domain randomization \cite{Tobin2017DomainRF} does not directly control worst‑case degradation, a DRO formulation emphasizes adverse extremes, thereby aligning with navigation safety.
% Data-Driven, a data-driven approach, defines the uncertainty set using the empirical distribution of the data, showing effectiveness in machine learning tasks such as regression, classification, and robust optimization \cite{staib2019distributionally}; \cite{kannan2020}.
% Several studies employ distributionally robust reinforcement learning to enhance robustness against historical data distribution shifts across environments and improve sample efficiency in RL.
Several studies employ DRO to enhance model robustness. However, these methods exhibit limitations: some necessitate auxiliary components (e.g., generative models \cite{shi2024curious, ramesh2024distributionally}) that complicate deployment; others are model-based \cite{shi2024distributionally, hesample} or rely on offline data \cite{Leung2025DistributionallyRP}. Such drawbacks restrict their applicability in real-world settings where RL agents must interact with environments online without prior environmental information and modeling design.

\section{Problem Formulation}
An USV does not have full knowledge of the environment, but instead receives noisy partial observations of robot's surroundings constrained by the limited sensing range of its sensors. We formulate the USV navigation problem as a \emph{perturbed} Partially Observable Markov Decision Process $\widetilde{\mathcal{M}}$, denoted as $(\mathcal{S}, \mathcal{A}, \mathcal{O}, \mathcal{P}, R, \gamma, \nu)$, where $\mathcal{S}$, $\mathcal{A}$ and $\mathcal{O}$ represent the state, action and observation space, respectively. 
At each time step $t$, the agent receives a corrupted observation $\nu(o_t)$ subject to natural disturbances, 
where $o_t \in \mathcal{O}$ denotes the observation of current state $s_t$. 
Based on this observation, the agent selects an action $a_t \in \mathcal{A}$ according to policy $\tilde{\pi}(a_t \mid \nu(o_t))$.
This leads the agent to transition to the next state $s_{t+1} \sim \mathcal{P}(\cdot | s_t, a_t)$, and the agent receives a reward $r_{t} \sim R(s_{t}, a_{t})$, as well as a perturbed observation $ \nu(o_{t+1})$ of $s_{t+1}$. 
Following observation-disturbed policy $\tilde{\pi}$, the discounted sum of future rewards is represented as the random variable $Z^{\tilde{\pi}}(s_t,a_t)$, defined as $ Z^{\tilde{\pi}}(s_t, a_t) = \sum_{k=0}^{\infty} \gamma^k R(s_{t+k}, a_{t+k})$, where $\gamma \in (0,1)$ denotes the discount factors and determines the relative importance of future rewards. 
The objective of standard RL under noisy observations is to maximize the expected discounted return, formalized by the value function: \( Q^{\tilde{\pi}}(s_t, a_t) = \mathbb{E}\left[Z^{\tilde{\pi}}(s_t, a_t)\right].\)
Expectations are risk-neutral, representing returns solely through $ Q^{\tilde{\pi}}(s,a) $.
DistRL addresses the limited expressiveness of this expectation by estimating the distribution of $ Z^{\tilde{\pi}}(s,a) $.
% Environmental noise is incorporated within the state transition kernel, effectively augmenting the state representation. This formulation preserves monotonicity and $\gamma$-shift invariance of the Bellman operator that ensures the value function can be evaluated for any stationary policy.
Environmental noise incorporated into the state transition kernel augments the state representation. This formulation preserves the Bellman operator's monotonicity and $\gamma$-shift invariance, enabling value function evaluation for any stationary policy.

\section{Methodology}
In this section, we elaborate on our DRO-enhanced DistRL method, DRIQN. We start by revisiting the theoretical foundations of DistRL, then formalize the task environment, and finally elucidate the architecture of the DRIQN approach.
\subsection{Distributional Reinforcement Learning}
% $$ V_{k+1}(s) = \max_{a} \left[ R(s,a) + \gamma \sum_{s'} P(s'|s,a) V_k(s') \right] \eqno{(1)} $$

% $$ Q^{\pi}(s,a) = \mathbb{E}_{s' \sim P} \left[ R(s,a) + \gamma \mathbb{E}_{a' \sim \pi} Q^{\pi}(s',a') \right] \eqno{(2)} $$

% $$ V^{\pi}(s) = \sum_{a} \pi(a|s) \left[ R(s,a) + \gamma \sum_{s'} P(s'|s,a) V^{\pi}(s') \right] \eqno{(3)} $$
Distributional Reinforcement Learning explicitly models the return distribution $Z$ governed by the following \textit{distributional} Bellman equation and optimality equation:
\begin{equation}
\label{zdef}
    Z^\pi(s,a) \stackrel{D}{=} R(s,a) + \gamma Z^\pi(s', a'),
\end{equation}
\begin{equation}
\label{operator}
    \mathcal{T}Z^\pi(s,a) \stackrel{D}{=} R(s,a) + \gamma Z^\pi\left( s', \operatorname*{argmax}_{a'} \mathbb{E}[Z^\pi(s',a')] \right),
\end{equation}
where $\stackrel{D}{=}$ indicates equality in distribution, and $\mathcal{T}$ represents the \emph{distributional Bellman optimality operator}.

We utilize DistRL with IQN as a risk-sensitive path planner that has achieved promising performance in many tasks of robotics \cite{liu2023adaptive,lin2023robust,shi2024robust}. 
IQN utilizes the quantiles to provide a flexible representation of return distribution $Z$, denoted as $Z_\tau \colon \!\! {=} F^{-1}_Z(\tau)$, where $\tau \sim U([0,1])$ and $F^{-1}$ is inverse cumulative distribution function. To effectively capture risk-sensitive behaviors, IQN incorporates a distortion function $\beta$ to adjust the quantile distribution, emphasizing certain regions of the return distribution. Then the distorted expectation of $Z$ under $\beta : [0,1] \to [0,1]$ is defined by:
\begin{equation}\label{qtoz}
Q_\beta(s, a) = \mathbb{E}_{\tau \sim U([0,1])} [Z_{\beta(\tau)}(s, a)].
\end{equation}

% Distributional RL algorithms focus on the return distribution, which satisfies the distributional Bellman equation (Eq. (5)), where $Z^\pi(s, a)$ is a random variable that satisfies $Q^\pi(s, a) = \mathbb{E}[Z^\pi(s, a)]$. Similarly, the distributional Bellman optimality equation (Eq. (6)) is used in this case.

The standard deep $Q$-Learning framework trains a neural network-based state-action value function by iteratively minimizing a loss function, which is defined through the temporal difference (TD) error. The TD error is defined as:
% \begin{equation}\label{tderror}
% \delta = \left[ r + \gamma \max_{a' \in \mathcal{A}} Q(x_{t+1}, a') - Q(x_t, a_t) \right]^2
% \end{equation}
\begin{equation}\label{tderror0}
\delta = r + \gamma \max_{a'} Q(s', a') - Q(s, a).
\end{equation}

The IQN loss function integrates the sampled temporal difference error (Eq.~(\ref{tderror1})) with the \textit{Huber} quantile regression \cite{huber1992robust} (Eq.~(\ref{quantilehuberloss})) to optimize the quantile value network:
\begin{equation}\label{iqnloss}
\mathcal{L}_{IQN}(s, a, r, s') = \frac{1}{N'} \sum_{i=1}^N \sum_{j=1}^{N'} \rho_{\tau_i}^{\kappa}(\delta^{\tau_i, \tau'_j})
\end{equation}
where $N$ and $N'$ denote the counts of i.i.d. samples $\{\tau_i\}_{i=1}^N$ and $\{\tau'_j\}_{j=1}^{N'}$ from $\mathcal{U}([0, 1])$. The temporal difference error and Huber loss components are:
\begin{equation}\label{tderror1}
\delta^{\tau_i, \tau'_j} = r + \gamma Z_{\tau'_j}(s', a') - Z_{\tau_i}(s, a)
\end{equation}
\begin{equation}\label{quantilehuberloss}
\begin{split}
    \rho_{\tau}^{\kappa}(u) &= |\tau - \mathbb{I}_{\{ u < 0 \}}| \left( \frac{\mathcal{L}_{\kappa}(u)}{\kappa} \right) \\
    \mathcal{L}_{\kappa}(u) &= 
    \begin{cases} 
        \frac{1}{2} u^2, & |u| \leq \kappa \\
        \kappa ( |u| - \kappa/2 ), & \text{otherwise}
    \end{cases}
\end{split}
\end{equation}

This integrated loss formulation ensures convex optimization of quantile representations during gradient-based training. The corresponding policy approximation uses $K$ quantile samples $\tilde{\tau}_k \sim \mathcal{U}([0,1])$:
\begin{equation}\label{policy}
\tilde{\pi}_\beta(s) = \operatorname*{argmax}_a \frac{1}{K} \sum_{k=1}^K Z_{\beta(\tilde{\tau}_k)}(s, a)
\end{equation}
with actions selected through $\tilde{\pi}_\beta(s)$.

\subsection{USV Navigation under Observational Perturbation}
% We extend the simulated marine environment \cite{lin2023robust}, which includes static obstacles and Rankine vortex-induced flows \cite{acheson1990elementary}. 
% While maintaining the original reward function, action range, and sensor settings, the robot must reach the goal while handling observation noise, avoiding obstacles, and adapting to flow disturbances.
In this task, the vehicle must reach the goal while handling natural observation noise, avoiding obstacles, and adapting to flow disturbances. The key components of RL are designed as follows:

% \subsubsection{Reward}
\noindent\textbf{Reward.} 
To learn a policy that moves toward the goal while avoiding obstacles, the reward function is designed as: $ r_t = r_{\text{step}} + \alpha (d_{t-1} - d_t) + r_{\text{collision}} \mathbb{I}_{\text{collision}} + r_{\text{goal}} \mathbb{I}_{\text{goal}} $, where \( d_t \) represents the distance between the robot and the goal at time \( t \). 
The indicator function \( \mathbb{I}_{\text{collision}} \) (resp. \( \mathbb{I}_{\text{goal}} \)) takes the value of 1 if collision happends (the goal is reached, resp) at \( t \), and 0 otherwise. The components of reward are set as \( r_{\text{step}} = -1.0 \), \( r_{\text{collision}} = -50.0 \), \( r_{\text{goal}} = 100.0 \), and \( \alpha = 1.0 \).

\noindent\textbf{Observation.} 
Based on real-world device, the observation at time step 
$t$ is given by: $ O_t = (O_{\text{velocity}}, O_{\text{goal}}, O_{\text{LiDAR}})$, where $O_{\text{velocity}}$ denotes the robot’s seafloor-relative velocity. $O_{\text{goal}}$ represents the goal position in the robot’s frame. $O_{\text{LiDAR}}$ consists of LiDAR reflections, which provide information about detected obstacles.

\noindent\textbf{Action.} 
The robot's action at time step \( t \) is defined as: $ a_t = (a(t), w(t)) $, where \( a(t) \) represents the rate of change in velocity magnitude, and \( w(t) \) denotes the rate of change in velocity direction. The available action space is: \(a \in \{-0.4, 0.0, 0.4\}~m/s^2\), \(\quad w \in \{-0.52, 0.0, 0.52\}~rad/s \). The forward speed is constrained within \([0, v_{\max}]\).

\noindent\textbf{Observational Perturbation.} 
Since our objective is to model natural noise shaped by coupled climate, meteorological, illumination, and sea‑state effects instead of adversarial attack scene, we rely on broadly observed, non‑synthetic noise sources with specifications given in Section \emph{Experimental Settings}.

% To reflect the heterogeneous noise induced by coupled real‑world factors (weather, sea state, geographic locality), we instantiate multiple simulated noise subgroups for conditions.
% Considering the context of online learning, at each step, decisions are made based on noisy perceived states. The subsequent perceptual inputs exhibit homogeneous noise profiles at each adjacent step within an episode, representing current natural condition. This setting requires the robot to adjust its strategy to account for cumulative errors. Due to the complex noise pattern, the noise type can be changed in other episodes. In this scenario, the policy will be adapted to perception noise $\nu$, the original state observation $ s $ will be $ \nu(s) $, and corresponding policy is expressed by $ \tilde{\pi}_\beta(\nu(s)) $.

% figure 2
\begin{figure}[t]
\centering
% \vspace{0.15cm}
\includegraphics[width=0.98\linewidth]{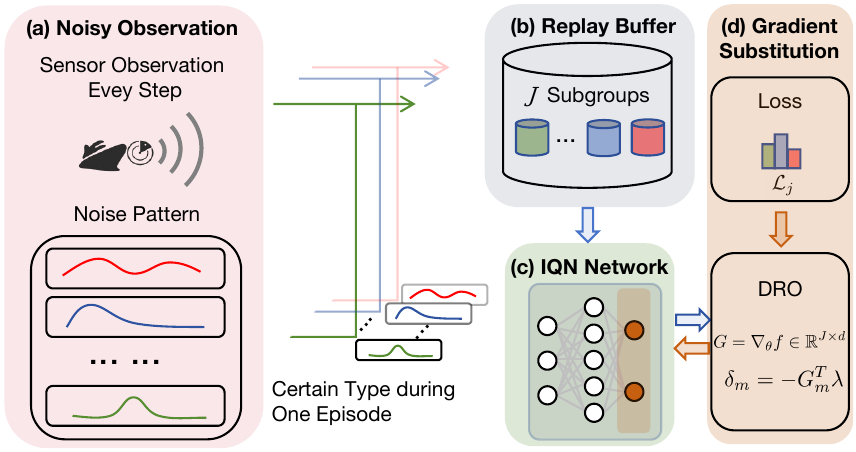}
\caption{\textbf{Overall framework of our method. }In (a), the USV receives noisy observations specific to its current natural condition, assuming a consistent noise pattern within one episode due to proximity of environmental condition. Therefore, the replay buffer in (b) accumulates subgroups with distinct noise patterns. This variation has a significant impact on the training of the implicit quantile networks in (c). To enhance robustness, DRIQN employs ``Gradient Substitution'' in (d), replacing the gradients of the output layer with those computed via DRO.}
\label{fig2}
\end{figure}
\subsection{Distributionally Robust Optimization for DistRL}
IQN has demonstrated superior performance as a risk-sensitive policy for path planning in an ideal environment. However, the planner relies on online distributional reinforcement learning training without prior information, meaning that errors in actions under noisy observations will cause reward signal mismatching and policy deviation. 
% This issue is further exacerbated by the presence of diverse noise types in the replay buffer, leading to the accumulation of contaminated experience data and compounding the challenge of learning a safe policy. 
This issue is exacerbated by the presence of multiple noise subgroups within the replay buffer.
Despite its impact, this problem remains largely unexplored in existing research.

% % Introduction
% The agent's perceptual system encounters distinct noise patterns across different regions of the environment during interactive episodes. As the agent sequentially engages with diverse environmental configurations, the accumulated experience replay data becomes contaminated with heterogeneous noise profiles. Crucially, each noise modality imposes unique biasing effects on policy optimization through its particular perturbation mechanisms. This multi-source noise interference induces compound distributional shifts in the replay buffer, where the mixture of contaminated data trajectories progressively diverges from the nominal state-action visitation distribution.

We first reformulate the algorithm's robustness facing noisy transition data as an optimization problem and then apply DRO to handle different patterns of noise that perceptions of robot sensors have. The framework of DRIQN is shown in Figure~\ref{fig2}.
% Without loss of generality, wethe noise patterns setting is a particular instance of Theorem 2 \cite{zhang2020robust} that satisfies the Bellman contraction. 

Based on previous optimization works about DRO \cite{Qi2020AnOM, shen2022towards}, the form of optimization target can be written as:
% \begin{equation}\label{droeq}
% \min_{\theta \in \mathbb{R}^d} \, F_p[\theta] = \min_{\theta \in \mathbb{R}^d} \, \sup_{p \in \mathbb{R}^n} \left\{ \sum_{i=1}^{n} p_i \mathcal{L}_{IQN}(g_{\theta}(x_i), y_i) - h(p, 1/n) \right\}
% \end{equation}
% \begin{equation}\label{eq:droeq}
% \min_{\theta \in \mathbb{R}^d} \, F_p(\theta) = \min_{\theta \in \mathbb{R}^d} \, \sup_{\mathbf{P} \in \mathbb{R}^n} \left\{ \sum_{i=1}^{n} p_i \mathcal{L}(h_{\theta}(T) - d(p, 1/n) \right\}
% \end{equation}
\begin{equation}\label{eq:droeq}
\min_{\mathbf{\theta} \in \mathbb{R}^d} \, F_p(\theta) = \min_{\mathbf{\theta} \in \mathbb{R}^d} \, \sup_{\mathbf{P} \in \mathbb{R}^n} \sum_{i=1}^{n} p_i \mathcal{L}_{IQN}(T_i) - d(\mathbf{P}, \mathbf{1}/n) ,
\end{equation}
where \(\mathcal{L}_{IQN}\) denotes the IQN loss function applied to the noisy transition datapoint \(T_i=(s_i,\,a_i,\,r_i,\,s_{i+1})\) from replay buffer of size \(n\). 
% The neural network parameter \(\mathbf{\theta} \in \mathbb{R}^d\) is associated with the IQN model, while \(\mathbf{P} = (p_1, p_2, \ldots, p_n)\) represents a vector of density ratios that characterizes the noise patterns. Additionally, the constraints \(\sum_{i=1}^{n} p_i = 1\) and \(p_i > 0\) hold. The term \(d(\mathbf{P}, \mathbf{1}/n)\) quantifies the difference between $\mathbf{P}$ and the uniform probabilities $\mathbf{1}/n$.
The IQN model uses parameters \(\mathbf{\theta} \in \mathbb{R}^d\), while \(\mathbf{P} = (p_1, \ldots, p_n)\) denotes noise-characterizing density ratios satisfying \(\sum p_i = 1\) and \(p_i > 0\). \(d(\mathbf{P}, \mathbf{1}/n)\) quantifies the difference between $\mathbf{P}$ and the uniform probabilities $\mathbf{1}/n$.
% The parameter \(\mathbf{\theta} \in \mathbb{R}^d\) parameterizes the IQN model, while \(\mathbf{P} = (p_1, p_2, \ldots, p_n)\) denotes a vector of density ratios characterizing noise patterns. These satisfy \(\sum_{i=1}^{n} p_i = 1\) and \(p_i > 0\). The term \(d(\mathbf{P}, \mathbf{1}/n)\) measures the divergence between \(\mathbf{P}\) and the uniform distribution \(\mathbf{1}/n\).

To adapt the DRO to big data and large parameter scenario, we focus on the uncertainty set $\mathcal{Q}$, which is partitioned into \(J\) subgroups that represent the subgroups of transition data with different noise types. Based on this replay buffer partition, we propose an efficient gradient-based learning approach to solve the optimization problem.

Given identical noise types per subgroup, we implement independent data sampling across all replay buffer subgroups.
To further improve both comprehensive robustness and data utilization efficiency beyond the worst-case subgroup, we propose computing the optimization gradient as a linear combination of the gradients from noisy subgroups.
The optimization gradient in DRIQN is used to replace the original gradient during the backpropagation through the last quantiles layer of the IQN network.

% We follow the simplified notation process of W-DRO \cite{shen2022towards}, the inner term of DRO (Eq. \eqref{eq:droeq}) will be:
We formalize the inner term of DRO problem Eq.~(\ref{eq:droeq}) with a partitioned uncertainty set $\mathcal{Q}$. For simplicity, we construct $\mathcal{Q}$ through considering limited $J$ kinds of noise, i.e. $\mathcal{Q}\colon \!\!\!\! =\{q_1,q_2,...,q_J\}$. 
\begin{equation}\label{eq:innerdroeq}
\sup_{\mathbf{P} \in \mathbb{R}^J, q \in \mathcal{Q}} \sum_{j=1}^{n} p_j \mathbb{E}_{(T_j) \sim q_j} \left[ \mathcal{L}_{IQN}(T_j) \right] - d(\mathbf{P}, \mathbf{1}/J).
\end{equation}
To ensure a fair comparison of natural-condition-induced noise types, we assume equiprobable occurrence for all categories. Formally, we define $\mathbf{P}$ as a uniform probability vector, which yields $d(\mathbf{P}, \mathbf{1}/J) = 0$ in the final term.
So the inner term of DRO will be expressed as:
\begin{equation}\label{eq:innerdroeq1}
\sup_{q \in \mathcal{Q}}\mathbb{E}_{T \sim q} \left[ \mathcal{L}_{{IQN}_{q}}(T) \right].
\end{equation}

Now we go back to the complete DRO minimax term Eq.~(\ref{eq:droeq}). To solve the problem, we aim to minimize the expected loss over the uncertainty set $\mathcal{Q}$ following:
\begin{equation}\label{eq:wholeopt}
\min_{\mathbf{\theta}} \max_{1 \leq j \leq J}
\frac{1}{c_j} \sum_{k=1}^{c_j} \mathcal{L}_{IQN}(T_{jk}),
\end{equation}
where \(J\) is the number of noise subgroups, \(c_j\) is the number of samples in subgroup \(j\) that has unique noise type, \(T_{jk}\) refers to the \(k\)-th transition data in the \(j\)-th subgroup. Our method provides a more scalable and efficient way to obtain the gradient descent direction by solving optimization problem Eq.~(\ref{eq:wholeopt}) through learning the supremum of the set of loss functions with respect to each noise subgroup.

As the optimization problem described in Eq.~(\ref{eq:wholeopt}), the right side noted as $f_j (\mathbf{\theta}) = \frac{1}{c_j} \sum_{k=1}^{c_j} \mathcal{L}_{IQN}(T_{jk})$ during the iteration $m$ can be linearized to achieve convex approximation, therefore we rewrite the right side as $f_j(\theta_m) + \langle \nabla f_j(\theta_m), \theta - \theta_m \rangle$. 

Since the inner term for $\max$ operation has been smoothed by linearization, to ensure finding the stable descent direction during the $\min$ stage when $\max_{1 \leq j \leq J}
f_j(\mathbf{\theta}_m) + \langle \nabla f_j(\mathbf{\theta}_m), \mathbf{\theta} - \mathbf{\theta}_m \rangle$ is not strictly convex with respect to parameter $\mathbf{\theta}$. Thus, we add a regularization term \( \| \theta - \theta_m \|_2 \). The gradient descent direction is set as \( \delta \colon\!\!\!\!= \theta - \theta_m \), we obtain the equivalent form of the DRO problem:
% \[
% \min_{\delta, \nu} \| \delta \|_a + \nu \tag{14a},
% \]
% \[
% \text{s.t. } f_j(\theta_m) + \langle \nabla f_j(\theta_m), \delta \rangle \leq \nu, \ \forall 1 \leq j \leq J. \tag{14b}
% \]
\begin{subequations}\label{eq:quadratic program}
\begin{align}
&\min_{\delta, \kappa} \| \delta \|_2 + \kappa \label{eq:main} \tag{13a}, \\
&\text{s.t. } f_j(\theta_m) + \langle \nabla f_j(\theta_m), \delta \rangle \leq \kappa, \ \forall 1 \leq j \leq J \label{eq:constraint} \tag{13b}.
\end{align}
\end{subequations}
% Then we apply the Theorem III.1. of W-DRO \cite{shen2022towards} to approximate supremum loss considering the dual problem of the semi-definite quadratic programming problem in \cref{eq:main,eq:constraint}. 

\begin{theorem}[Equivalent dual quadratic program]
\itshape % set tilt
\label{thm:Equivalent Dual Quadratic Program}
Let $\theta_m\in\mathbb{R}^{d}$ be the model parameter at iteration $m$ and define the subgroup losses 
$
    f_j(\theta)\,,\; j=1,\dots,J.
$
Denote $G_m \colon \!\! {=} \nabla_\theta f(\theta_m)\in\mathbb{R}^{J\times d}$, whose $j$‑th row is $\nabla_\theta f_j(\theta_m)^{\!\top}$,  
and consider the quadratic program in Eq.~(\ref{eq:quadratic program}).
% \begin{equation}
% \label{eq:primal}
% \begin{aligned}
% \min_{\delta\in\mathbb{R}^d,\;\nu\in\mathbb{R}}\;&
%         \tfrac{1}{2}\|\delta\|_2^{2}+\nu \\
% \text{s.\,t.}\quad &
%         f_j(\theta_m)+\langle\nabla_\theta f_j(\theta_m),\delta\rangle\le \nu,\quad  j=1,\dots,J .
% \end{aligned}
% \end{equation}
The unique optimal descent direction of this problem is a convex combination form:
\begin{equation}
\delta_m^\star = -G_m^{\top}\lambda^\star, \quad 
\lambda^\star \in \Delta_J \colon \!\! {=} 
\Bigl\{ \lambda \in \mathbb{R}_{\ge 0}^{J} \Bigm| \sum_{j=1}^{J} \lambda_j = 1 \Bigr\},
\end{equation}
where $\lambda^\star$ solves the dual quadratic program
\begin{equation}
\label{eq:dual}
\min_{\lambda\in\Delta_J}\;
        \tfrac{1}{2}\,\lambda^{\top}G_mG_m^{\top}\lambda-\lambda^{\top}f(\theta_m).
\end{equation}
Hence $\delta_m^\star$ is a linear combination of all subgroup gradients at iteration~$m$.

\end{theorem}

The proof of Theorem 1 following previous work \cite{shen2022towards} is given in Appendix A. This theorem indicates that the original constrained min-max problem can be reformulated as a dual quadratic programming problem with dimensionality $J$ (the number of noise subgroups). Instead of optimizing high-dimensional vectors, the solution now consists of a linear combination of subgroup gradients. This reformulation reduces the computational cost to $O(J^3)$.
Statistically, the convex combination coefficients serve as a weight vector assigning importance across subgroups. Theorem 1 proves that DRIQN can enhance data utilization compared to methods relying solely on worst-case scenarios.

For the convergence of DRIQN, the model training is based on the stochastic approximation of the gradient. Due to the boundedness of the mini-batch gradients of the objective function $\max_{1 \leq j \leq J} f_j(\mathbf{\theta}_m) + \langle \nabla f_j(\mathbf{\theta}_m), \mathbf{\theta} - \mathbf{\theta}_m \rangle$, the stochastic convergence rate is $\mathcal{O}(T^{-1/2})$ following the conclusion in \cite{Hazan2006LogarithmicRA}, where $ T $ denotes the iteration times.

\begin{figure*}[t]
  \centering
% \vspace{0.3cm}
\includegraphics[width=0.99\linewidth]{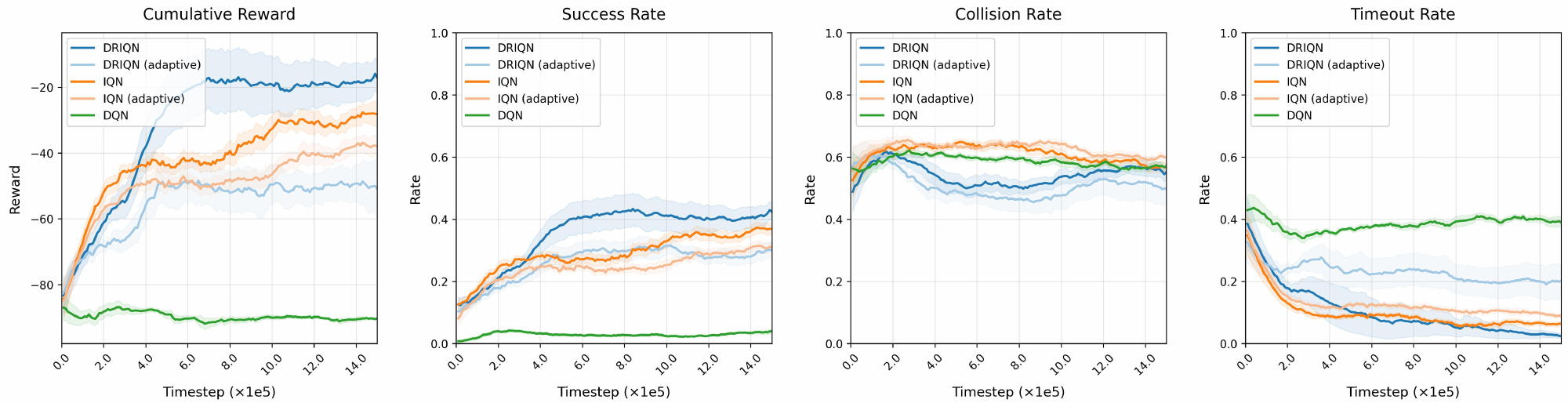}
  \caption{Online evaluations compare three learning-based models and two distributional RL strategies. Performance, safety, and convergence are assessed via three key metrics with cumulative reward curves. Implementation details appear in \emph{Experimental Settings}. Due to space limit, the other two figures beyond noise level 0.6 are in Appendix B.}
  \label{fig:exp1}
\end{figure*}

\section{Experiments}\label{secv}

% REATTAIN-IQN(data-dRivEn distributionAlly robusT opTimizAtion wIth iqN) / DDDRO-IQN

\begin{table*}[ht]
% \small
\footnotesize
\centering
% \small
% \footnotesize
\setlength{\tabcolsep}{0.34mm}
% \renewcommand{\arraystretch}{1}
% \newcolumntype{C}[1]{>{\centering\arraybackslash}p{#1}}
% \resizebox{\textwidth}{!}{%
% \begin{tabular}{C{2.3cm}|C{1.6cm}|ccc|ccc|ccc|ccc|ccc|ccc}
\begin{tabular}{c|c|ccc|ccc|ccc|ccc|ccc|ccc}
\toprule
% \multirow{2}{*}{\diagbox[width=2.3cm, height=0.85cm]{\small\textbf{Method}}{\small\textbf{Metrics}}} &
\multirow{2}{*}{{\small\textbf{Method}}} &
\multirow{2}{*}{\textbf{Strategy}} &
\multicolumn{3}{c|}{\textbf{Success Rate} $\uparrow$} &
\multicolumn{3}{c|}{\textbf{Collision Rate} $\downarrow$} &
\multicolumn{3}{c|}{\textbf{Timeout Rate} $\downarrow$} &
\multicolumn{3}{c|}{\textbf{Final Reward} $\uparrow$} &
\multicolumn{3}{c|}{\textbf{Avg Time} $\downarrow$} &
\multicolumn{3}{c}{\textbf{Avg Energy} $\downarrow$} \\
\cmidrule(lr){3-20}
& &
\textbf{0.6} & \textbf{0.4} & \textbf{0.2} &
\textbf{0.6} & \textbf{0.4} & \textbf{0.2} &
\textbf{0.6} & \textbf{0.4} & \textbf{0.2} &
\textbf{0.6} & \textbf{0.4} & \textbf{0.2} &
\textbf{0.6} & \textbf{0.4} & \textbf{0.2} &
\textbf{0.6} & \textbf{0.4} & \textbf{0.2} \\
\midrule
APF   & --       & 0.11 & 0.25 & 0.03 & 0.81 & 0.55 & 0.53 & 0.08 & \underline{0.20} & 0.44 & -- & -- & -- & 206.68 & 349.32 & 547.89 & 320.96 & 654.48 & 1071.70 \\
\midrule
BA    & --       & 0.22 & 0.11 & 0.26 & 0.71 & 0.46 & \textbf{0.33} & 0.07 & 0.43 & 0.41 & -- & -- & -- & 185.76 & 482.98 & 462.97 & 251.11 & 691.93 & 691.57 \\
\midrule
DQN   & Greedy   & 0.04 & 0.03 & 0.02 & \underline{0.57} & 0.57 & 0.57 & 0.39 & 0.40 & 0.41 & -90.30 & -90.11 & -89.54 & 541.27 & 550.35 & 560.52 & 858.39 & 779.78 & 832.67  \\
\midrule
\multirow{2}{*}{IQN} 
      & Greedy   & \underline{0.37} & \underline{0.50} & \underline{0.61} & \underline{0.57} & \underline{0.45} & \underline{0.37} & \underline{0.06} & \textbf{0.04} & \underline{0.02} & \underline{-28.16} & \underline{-14.87} & \underline{13.47} & \underline{144.89} & \underline{123.75} & \underline{61.25} & \underline{196.73} & \underline{181.63} & \underline{92.18} \\
      & Adaptive & 0.31 & 0.36 & 0.46 & 0.60 & 0.51 & 0.38 & 0.09 & 0.13 & 0.16 & -37.93 & -38.97 & -48.66 & 185.75 & 230.00 & 274.27 & 247.34 & 296.00 & 358.99 \\
\midrule
\multirow{2}{*}{DRIQN}
      & Greedy   & \textbf{0.42} & \textbf{0.55} & \textbf{0.63} & 0.55 & \textbf{0.42} & \underline{0.37} & \textbf{0.02} & \textbf{0.04} & \textbf{0.01} & \textbf{-17.11} & \textbf{3.14} & \textbf{26.87} & \textbf{93.51} & \textbf{90.11} & \textbf{31.82} & \textbf{141.67} & \textbf{114.15} & \textbf{53.61} \\
      & Adaptive & 0.30 & 0.39 & 0.33 & \textbf{0.50} & 0.46 & 0.49 & 0.20 & 0.15 & 0.18 & -50.84 & -36.16 & -45.90 & 283.85 & 226.07 & 272.25 & 392.90 & 257.17 & 401.26 \\
\midrule
\multicolumn{2}{c|}{\textbf{Improvement (\%)}} &
\textbf{13.51} & \textbf{10.00} & \textbf{3.28} & \textbf{12.28} & \textbf{6.67} & -12.12 & \textbf{66.67} & -- & \textbf{50.00} & \textbf{39.24} & \textbf{121.12} & \textbf{99.48} & \textbf{35.46} & \textbf{27.18} & \textbf{48.05} & \textbf{27.99} & \textbf{37.15} & \textbf{41.84} \\
\bottomrule
\end{tabular}
\caption{\textbf{Overall performance comparison} under three different noise levels (0.6, 0.4, and 0.2) related to the variances of the noise distributions. The best results are highlighted in \textbf{bold} and the second best results are highlighted with an \underline{underline}. ``Improvement'' denotes the percentage increase (positive metrics) or decrease (negative metrics) over the second best models.}
\label{tab:standard_metrics}
\end{table*}

\subsection{Experimental Settings}

\noindent \textbf{Simulation Environment.} 
To bring the simulation closer to real sea-surface conditions characterized by uncertain hazards such as inconspicuous current vortices and reef obstacles, we employ a risky marine environment under perturbation based on previous work \cite{lin2023robust} and deploy a purely sensor-based navigation scheme without any prior knowledge. 
% As this work investigates robust policy optimization under severely noise‑degraded observations, we adopt the environment complexity with four vortices and six obstacles. 
% To investigate robust policy optimization under severely noise-degraded observations, we utilize a complex environment comprising four vortices and six obstacles.
% Since this work addresses robust policy optimization in the presence of severely noise-degraded observations, we implement a risky marine environment under perturbation, including  four vortices and six obstacles.
Given our focus on robust policy optimization under severe observational noise, the environment features 4 vortices and 6 obstacles randomly dispersed across the operational space.
To mitigate randomness, we train the models for a total of 1.5 million time steps across nine different random seeds.
During the evaluation stage, the positions of both the goal and start point are reset each time, allowing exploration of different environmental conditions and capturing the varying noise patterns.
Agents undergo evaluation every 10,000 training steps across 15 pre-randomized environments.
% Every agent is evaluated on 15 pre-randomly generated environments after every 10,000 steps of training across nine different random seeds to mitigate randomness. 
% To ensure fairness in the results, we use the hyperparameter settings of RL methods from previous work \cite{lin2023robust}.

\noindent \textbf{Noise Settings.} 
To enhance the perception robustness of IQN agent for our purposes, we implement the DRIQN based on this environment and posit that the critical challenge lies in maintaining the robustness of DistRL under noisy observation.
In order to narrow the gap between simulation and real sea-surface sensing, we ground our noise modeling in prior physical studies \cite{gazeboKoenigHoward2004, Laconte2021DynamicLA} and superimpose ubiquitous non‑synthetic sensor noise processes, specifically Gaussian and Poisson noise. 
To assess DRIQN's multi-noise robustness, \emph{The Analysis of ``Noise-Type''} incorporates extra salt-and-pepper noise (simulating extreme value fluctuations) and occlusion noise (emulating partial sensor failures).
This DRO framework is agnostic to the specific noise law, enabling extensions to other settings, including safety analysis under adversarial perturbations. 
The noise intensity is configured as a tunable hyperparameter, with values empirically set to \{0.6, 0.4, 0.2\} to evaluate performance degradation across perturbation levels systematically.
The sensor value range and noise intensity both determine the distributions. 

\noindent \textbf{Baselines.}
We compare DRIQN with the following representative methods in USV navigation:
(1) Classical observation-based planning methods: Artificial Potential Field (APF) method \cite{Fan2020ImprovedAP} and Bug Algorithm (BA) method \cite{Lumelsky1988APF}; (2) Standard Q-learning method in DRL: DQN \cite{Mnih2015HumanlevelCT}; (3) Risk-sensitive method in DistRL: IQN \cite{dabney2018implicit}.

\noindent \textbf{Hyperparameter Settings.} 
For neural networks with three or more layers, we implement gradient substitution exclusively in the final layer instead of the whole layers, inspired by established computer vision research on gradient manipulation \cite{shen2022towards}. The comparative analysis is detailed in Section: \emph{The Analysis of ``Gradient Substitution''}.
Additionally, to adapt the gradients computed by the DRO to the training of all layers except the final one, we adapt the learning rate decay of these layers starting from 0.0001 to 0.000001 and appropriately shrink the calculated gradients. 
The experiments were completed in parallel using six A6000 GPUs and AMD EPYC 7542 CPU.

\subsection{Metrics and Strategy}\label{ss:Metrics and Strategy}

% \begin{figure*}[t]
%   \centering
%   \vspace{0.2cm}
%   \includegraphics[width=0.98\linewidth]{exp5_h.pdf} % 关键修改：width=\columnwidth
%   % \vspace{-0.3cm}
%   \caption{\textbf{Qualitative trajectory results} (noise variance level = 0.6). The greedy approach avoids risk-sensitive adjustments triggered by noisy observations, unlike the adaptive strategy. Aligning with Table \ref{tab:standard_metrics}, the adaptive strategy exhibits higher timeout rates and frequent deviations from complex regions to mitigate risk, as seen in repeated trials.}
%   \vspace{-0.3cm}
%   \label{fig:exp5}
% \end{figure*}

\noindent \textbf{Metrics.} 
To evaluate the performance of our approach, we employed six metrics encompassing performance, safety, and efficiency: success rate (SR), collision rate (CR), timeout rate (TR), final (cumulative) reward (FCR), average time consumed (AT), and average energy consumed (AE). Average time and energy are calculated based solely on data from successful episodes.
% The inherent randomness in RL training process causes significant fluctuations in the original evaluation curves. 
% To clearly illustrate the trend of metric changes during training, this study applies an Exponential Moving Average (EMA) to smooth the raw evaluation data. Following previous work \cite{henderson2018deep}, a smoothing factor of 25 is chosen in this work. 

\noindent \textbf{Strategy.} 
We compare the standard greedy strategy, which always selects the action with the highest expected reward, with an adaptive strategy for the IQN-based methods proposed by \cite{lin2023robust}. The adaptive strategy adaptively tailors the region of interest in return distributions through computing CVaR threshold based on real-time environmental risk assessments. This threshold reflects different levels of risk sensitivity, measured by the robot’s distance to the nearest obstacles. When no obstacles are detected, the adaptive strategy behaves like a greedy one.

% \begin{figure}[!t]
%   \centering
%   \vspace{0.2cm}
%   \includegraphics[width=0.82\columnwidth]{exp_2c.pdf} % 关键修改：width=\columnwidth
%   % \vspace{-0.3cm}
%   \caption{\textbf{Trajectory performance comparison. } Compared to the adaptive strategy, the greedy strategy avoids unnecessary noise based risk-sensitive adaptations in noisy observations. Consistent with the results in Table \ref{tab:standard_metrics}, the adaptive strategy exhibits a higher timeout rate and a tendency to deviate from designated complex regions to avoid risk, as observed across multiple trajectories. Due to the risk-sensitive nature, adaptive strategy struggles to make effective decisions under noisy conditions, relying heavily on sensor accuracy.}
%   \vspace{-0.3cm}
%   \label{fig:exp5}
% \end{figure}

\begin{figure*}[t]
  \centering
  % \vspace{0.2cm}
  \includegraphics[width=0.99\linewidth]{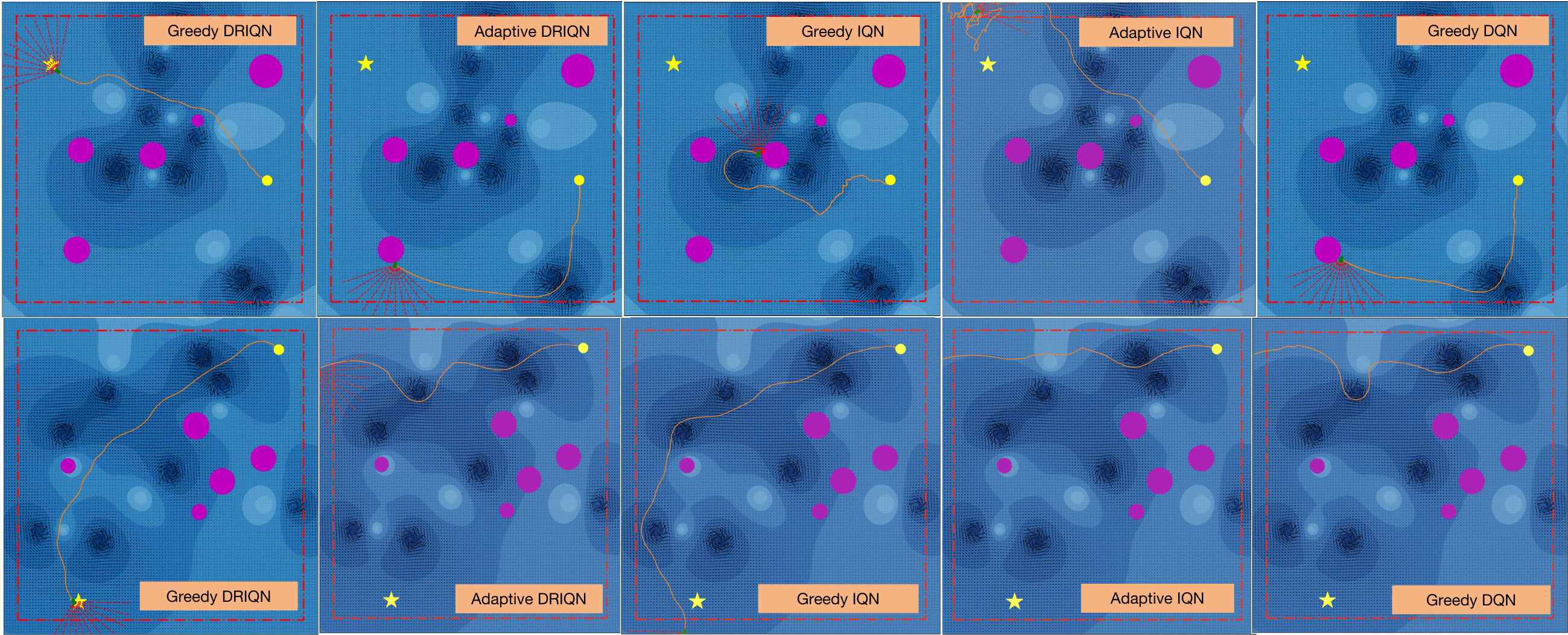} % 关键修改：width=\columnwidth 以及 \vspace{-0.3cm}
  \caption{\textbf{Qualitative trajectory results} (noise variance level = 0.6). The greedy approach avoids risk-sensitive adjustments triggered by noisy observations, unlike the adaptive strategy. Aligning with Table~\ref{tab:standard_metrics}, the adaptive strategy exhibits higher timeout rates and frequent deviations from complex conditions to mitigate risk, as seen in repeated trials.}
  \label{fig:exp5}
\end{figure*}

\subsection{Performance Comparison}

Online evaluation results displayed in Figure~\ref{fig:exp1} are smoothed by Exponential Moving Average to describe learning curves with shaded regions showing standard deviations over 9 seeds under high noise variance (level=0.6).
% Complete results appear in Appendix B.

Key performance illustrated in Table~\ref{tab:standard_metrics} are further summarized as follows:
\begin{itemize}
    \item \textit{DRIQN vs. Baseline Methods}: As a DistRL approach enhanced by DRO-based gradient substitution, DRIQN achieves superior success / collision rates and impressive efficiency compared with other traditional physical methods and RL algorithms across noise intensities, particularly under high noise variance. 
    \item \textit{DistRL vs. Standard RL}: DistRL methods consistently outperform standard deep Q-learning. DQN suffers from high timeout rates and excessive resource consumption due to passive collision avoidance behaviors, a limitation rooted in standard RL's dependence on expected returns that provide inadequate information in noisy environments. In contrast, DistRL's explicit modeling of return distributions yields inherent noise robustness.
    \item \textit{Greedy vs. Adaptive}: Both IQN and DRIQN show higher success rates with greedy action selection. Adaptive strategies suffer under noise interference as perturbed value distributions impair their decision mechanism.
    \item \textit{DRIQN Efficiency}: Under stringent noise-corrupted conditions, greedy DRIQN maintains the lowest timeout rate while reducing energy and time consumption, optimally balancing safety and efficiency. 
\end{itemize}

The qualitative trajectory results of RL agents are visualized in Figure~\ref{fig:exp5}.

\begin{table}[t]
\centering
\small
\begin{tabular*}{\columnwidth}{@{\extracolsep{\fill}}lcccccc@{}}
\toprule
Model & SR $\uparrow$ & CR $\downarrow$ & TR $\downarrow$ & FCR $\uparrow$ & AT $\downarrow$ & AE $\downarrow$ \\
\midrule
APF   & $0.12$ & $0.80$ & $0.08$ & --      & $211.83$ & $351.38$ \\
BA    & $0.14$ & $0.82$ & $0.04$ & --      & $134.65$ & $198.20$ \\
DQN   & $0.11$ & $0.50$ & $0.39$ & $-80.02$& $507.04$ & $798.70$ \\
IQN   & $0.54$ & $0.44$ & $0.02$ & $-3.32$ & $75.27$  & $108.54$ \\
DRIQN & $\mathbf{0.58}$ & $\mathbf{0.41}$ & $\mathbf{0.01}$ & $\mathbf{12.45}$ & $\mathbf{45.10}$ & $\mathbf{73.30}$ \\
\bottomrule
\end{tabular*}
\caption{Comparison of multi-noise robustness performance, with the best method highlighted in \textbf{bold}.}
\label{tab:4subgroups}
\end{table}

\subsection{The Analysis of ``Noise Types''}
To rigorously evaluate DRIQN's safety in multi-noise conditions, we augment natural noise conditions with salt-and-pepper noise (simulating sensor malfunction) and occlusion noise (emulating data loss scenarios) for faulty sensors, setting a fractionally lower noise intensity across all sufficient perturbations for intensity-diversity balance. The strategy is rigged to greedy rather than adaptive for more competitive performance in this task, as empirically demonstrated earlier. 
% Evidenced in Table \ref{tab:4subgroups}, DRIQN's optimization across heterogeneous noise subgroups shows superior capability in handling complex perceptual patterns and corresponding observational robustness compared to baseline methods.
Evidenced in Table~\ref{tab:4subgroups}, DRIQN's optimization across heterogeneous noise subgroups demonstrates superior handling of complex perceptual patterns and observational robustness over baselines.

% \begin{table}[h]
% \centering
% \small
% % \setlength{\tabcolsep}{1.0 pt} % 调整列宽
% % \renewcommand{\arraystretch}{1.2}
% \begin{tabular}{@{}lcccc@{}}
% \toprule
% \textbf{Model} & \textbf{Strategy} & \textbf{Success Rate} & \textbf{Collision Rate} & \textbf{Avg Time} \\
% \midrule
% \multirow{2}{*}{DRIQN}
%  & Greedy   & $\mathbf{0.42}$ & $0.55$ & $\mathbf{93.51}$ \\
%  & Adaptive & $0.30$ & $\mathbf{0.50}$ & $283.85$ \\
% \midrule
% \multirow{2}{*}{DRIQN-W}
%  & Greedy   & $\underline{0.22}$ & $\underline{0.54}$ & $\underline{357.87}$ \\
%  & Adaptive & $0.20 $ & $0.55$ & $379.50$ \\
% \bottomrule
% \end{tabular}
% \caption{Architectural comparison of ``gradient substitution'' under high noise variance: Whole-network substitution (DRIQN-W) versus final-layer substitution (DRIQN).}
% \label{tab:suball}
% \end{table}

\begin{table}[t]
\centering
\small
\begin{tabular*}{\columnwidth}{@{\extracolsep{\fill}}lcccc@{}}
\toprule
Model & Strategy & SR $\uparrow$ & CR $\downarrow$ & AT $\downarrow$ \\
\midrule
\multirow{2}{*}{DRIQN}
 & Greedy   & $\mathbf{0.42}$ & $0.55$ & $\mathbf{93.51}$ \\
 & Adaptive & $0.30$ & $\mathbf{0.50}$ & $283.85$ \\
\midrule
\multirow{2}{*}{DRIQN-W}
 & Greedy   & $\underline{0.22}$ & $\underline{0.54}$ & $\underline{357.87}$ \\
 & Adaptive & $0.20$ & $0.55$ & $379.50$ \\
\bottomrule
\end{tabular*}
\caption{Architectural comparison of ``gradient substitution'' under high noise variance: Whole-network substitution (DRIQN-W) versus final-layer substitution (DRIQN).}
\label{tab:suball}
\end{table}

\subsection{The Analysis of ``Gradient Substitution''}
% To investigate the design of last-layer substitution in DRIQN, we introduce the variant DRIQN-W that utilizes whole-network (4 layers) gradient substitution. The comparison in Table \ref{tab:suball} shows that the convenient last-layer substitution achieves both computational efficiency and superior performance by leveraging the deep network's representational capacity without destabilizing foundational features, resulting in better convergence properties compared to full-network gradient modification.
% To examine last-layer substitution in DRIQN, we introduce variant DRIQN-W using whole-network (4 layers) gradient replacement. Table \ref{tab:suball} demonstrates the approach with last-layer substitution achieves both computational efficiency and superior performance since it leveraging the deep neural network's representational capacity while preserving foundational features even under the most challenging regime for robustness.
To examine our last-layer gradient substitution setting for DRIQN, we introduce the variant DRIQN-W featuring whole-network (4-layer) gradient replacement for comparison. 
% As evidenced in Table~\ref{tab:suball}, the last-layer substitution setting retains the representational advantages of deep neural networks while enhancing robustness through streamlined optimization, achieving superior performance and computational efficiency even under extreme noise conditions.
As evidenced in Table~\ref{tab:suball}, last-layer substitution significantly outperforms whole-network gradient replacement in both success rate and computational efficiency while ensuring better safety. This advantage stems from its capacity to retain deep neural networks' representational benefits while enhancing robustness through streamlined optimization, even under extreme noise conditions. Learning curves comparison in Appendix C shows that last-layer substitution has better convergence than full-network modification.

\section{Conclusion}

In this paper, we introduce DRIQN, a robust and efficient distributional RL policy for USV path planning in unknown marine environments subject to observation noise. DRIQN addresses limitations of vanilla DistRL with implicit quantile networks in robustness and efficiency by employing a flexible gradient substitution method that uses DRO for gradient calculation, enabling the policy to handle complex observation noise patterns.
% Experimental results show that the DRIQN achieves superior performance in all six metrics encompassing performance, safety and efficiency, outperforming the state-of-the-art methods. 
Experimental results confirm DRIQN’s superiority across all six metrics in performance, safety, and efficiency domains, outperforming SOTA methods.
% In addition, we found that the greedy strategy outperforms the adaptive strategy of DistRL under noisy observations.
Results also reveal that within DistRL, the greedy strategy outperforms adaptive counterpart under noisy observational conditions.
In future work, we plan to explore additional DistRL methods to enhance robustness and broaden our approach to more diverse environmental settings.
We also aim to validate effectiveness of these methods in real-world deployments, ensuring practical applicability.

\bibliography{aaai2026}

\end{document}